\documentclass[letterpaper, 10 pt, conference]{ieeeconf}  
\usepackage{graphicx} 
\usepackage[font=small,labelfont=bf]{caption} 
\usepackage{amsmath} 
\usepackage{mathtools}
\usepackage{bm}
\usepackage{amssymb}
\usepackage{algorithm} 
\usepackage{algpseudocode} 
\usepackage{cite}
\usepackage{svg}
\usepackage{booktabs} 
\usepackage{multirow} 

\DeclareMathOperator*{\argmin}{arg\,min}

\IEEEoverridecommandlockouts                              

\title{\LARGE \bf
Integrating Higher-Order Dynamics and Roadway-Compliance into Constrained ILQR-based Trajectory Planning for Autonomous Vehicles
}

\author{Hanxiang Li$^{*}$, Jiaqiao Zhang$^{*}$, Sheng Zhu$^{*}$, Dongjian Tang$^{*}$, Donghao Xu$^{*}$ 
\thanks{*The authors are with Deeproute.ai Ltd., 3125 Skyway Ct, Fremont, CA 94539 USA (email: hxlee@umich.edu, jiaqiao.zhang@alumni.stanford.edu, zhu.1473@osu.edu, tangdongjian@hotmail.com, donghaoxu@deeproute.ai)}
}

\begin{document}

\maketitle
\thispagestyle{empty}
\pagestyle{empty}

\begin{abstract}


This paper addresses the advancements in on-road trajectory planning for Autonomous Passenger Vehicles (APV). Trajectory planning aims to produce a globally optimal route for APVs, considering various factors such as vehicle dynamics, constraints, and detected obstacles. Traditional techniques involve a combination of sampling methods followed by optimization algorithms, where the former ensures global awareness and the latter refines for local optima. Notably, the  Constrained Iterative Linear Quadratic Regulator (CILQR) optimization algorithm has recently emerged, adapted for APV systems, emphasizing improved safety and comfort. However, existing implementations utilizing the vehicle bicycle kinematic model may not guarantee controllable trajectories. We augment this model by incorporating higher-order terms, including the first and second-order derivatives of curvature and longitudinal jerk. This inclusion facilitates a richer representation in our cost and constraint design. We also address roadway compliance, emphasizing adherence to lane boundaries and directions, which past work often overlooked. Lastly, we adopt a relaxed logarithmic barrier function to address the CILQR's dependency on feasible initial trajectories. The proposed methodology is then validated through simulation and real-world experiment driving scenes in real time. 

\end{abstract}

\section{INTRODUCTION}

The goal of on-road trajectory planner is to generate a globally optimal trajectory for APV, subjected to chosen vehicle dynamics, cost and constraint design. Input to the module includes detected obstacles' position, shape, and future predicted trajectories, road structure, and high-level behavior decisions. One of the primary methodologies employed for on-road trajectory planning involves the utilization of sampling techniques coupled with optimization algorithms \cite{phdthesis}\cite{li2022autonomous}. Sampling-based planner provides good global awareness and generates a resolution-complete trajectory, which can be used as the nominal trajectory for the follow-up optimization stage to further generate the locally optimal trajectory.

CILQR, an optimization algorithm, has been recently proposed and adapted to the APV system to enhance its overall comfort and safety performance. Previous work \cite{8671755}\cite{ma2022alternating} applied a vehicle bicycle kinematic model but failed to capture higher-order physical terms, and therefore does not guarantee the controllability of the trajectory. We incorporate higher order terms into the vehicle model, including the first and second order derivatives of curvature, as well as longitudinal jerk. We then leverage these additional state and action terms in our cost and constraint design, with various case study demonstration.

In on-road trajectory planning, ensuring roadway compliance is crucial. This includes adhering to key guidelines such as staying within designated lane boundaries, following lane directions, and maintaining lane centering. Some previous work, however, merely introduces the reference tracking term to encourage APV to stay close to the lane center, without consideration of non-crossable lane boundary and lane direction, which may lead to roadway-noncompliance. We introduce additional cost and constraints to address such requirement. To deal with hard constraints, generally a barrier function such as the standard logarithmic function would be used in CILQR algorithm, but it would require the initial guess must be feasible. We adopted relaxed logarithmic barrier functions similar to \cite{shimizu2020motion} to address such issue. 


\section{RELATED WORK}

Autonomous driving research has been popular over the past ten years and different trajectory planning techniques have been introduced to solve the problem of autonomous driving under complex on-road scenarios \cite{article}. The prevailing trajectory planning methodologies can be classified into two principal categories: sampling-based algorithms and optimization-based algorithms. Sampling-based algorithms \cite{5980223} search in discretized space which can result in a non-smooth trajectory and is often followed by optimization-based smoothing algorithms. Optimization-based approach \cite{xiao2021bridging} formulates the trajectory planning problem as an optimization problem, which allows planning in continuous space compared to the sampling-based approach and can generate a smooth trajectory directly.

\subsection{Sampling-based Trajectory Planning Method}

Many sampling-based algorithms are formulated within a spatial-temporal framework, aimed at reducing the planning dimension to a constrained 3D optimization problem. \cite{5980223} proposes a state lattice framework that represents the search space as 3D state lattices, allowing for direct exploration of the 3D dimension to identify the optimal solution. \cite{5354448} further extends this approach with trajectory sampling, which increases the sampling efficiency. However, the computational load of these approaches with dense sample is still not acceptable on real-time planners due to search complexity. \cite{fan2018baidu} further partitioned the space-temporal domain into path planning and speed planning to reduce the computational load. However, it introduces discrepancies between path optimization and speed optimization and leads to suboptimal results.

\subsection{Optimization-based Trajectory Planning Method}

There are mainly two types of optimization-based methods: direct and indirect \cite{optimization}. The fundamental concept behind the direct method involves treating points along the trajectory as decision variables and optimizing the trajectory by directly modifying these trajectory points. The work by \cite{6856581} uses this method and solves the optimization problem by sequential quadratic programming (SQP) for the autonomous driving problem. Although the elimination of system dynamics improves the computation efficiency significantly, this approach still suffers from high computational load when applied in on-road scenarios with long planning horizon, complex non-convex constraints, and nonlinear system dynamics.

The indirect method basically uses control inputs as decision variables and obtains the optimal trajectory by forward propagation with the system dynamic equations. One example is Differential Dynamic Programming (DDP) \cite{Jacobson1968NewSA} which can solve unconstrained optimal control problems efficiently. Iterative Linear Quadratic Regulator (ILQR) \cite{1469949} simplifies the non-linear system approximation in DDP from second order to first order which allows faster computation. \cite{8671755} further explores considering constraints in ILQR but is only limited to comfort and safety constraints. \cite{shimizu2020motion} proposes using relaxed logarithmic barrier functions avoid infeasible initial trajectory issue in CILQR, while \cite{ma2022alternating} utilizes Alternating Direction Method of Multipliers (ADMM) to circumvent the feasibility requirement of the initial trajectory. We believe there is still a lack of methods that incorporate high-order system dynamics and comprehensive consideration of complex on-road scenario constraints, which are essential for real-world on-road APV application. 

\section{METHODOLOGY}

\subsection{Iterative Linear Quadratic Regulator}

For ILQR algorithm, it optimizes upon a nominal trajectory, which is denoted as $(\Tilde{x}, \Tilde{u})$. In the first iteration, it either comes from the previous executed trajectory or an upstream sample-based trajectory planner. The goal is to find the state and action increment sequence pair $(\delta x^*, \delta u^*)$ that locally optimizes the trajectory.

\subsubsection{Unconstrained Trajectory Optimization Problem}

\begin{equation} \label{unconstrained traj opt}
\begin{split}
x^{*}, u^{*} = \argmin_{x, u} [{\sum_{0}^{N-1} L^{k} (x_k, u_k)} +L_F(x_N)] \\ 
s.t. \: x_{k+1} = f(x_k, u_k), \: x_0 = x_{start} 
\end{split}
\end{equation}
where $x_k$ is the state vector and $u_k$ is the action vector in time step k, N is the number of planning steps, ${f(. , .)}$ is the vehicle model, ${L^{k}(. , .)}$ is the k-th step cost, $L_F(.)$ is the final state cost, $x_{start}$ is the planning start point, $x^*$ and $u^*$ are the states and actions of the locally optimal trajectory. \\

\subsubsection{Backward Pass} \hfill\\

The goal of this step is to solve for the optimal action increment sequence in a backward order.

$V^k$ is the optimal cost-to-go start at k-th step ($k \in [0, N]$), and its definition is given below:
\begin{align}
V^N (x_N) &= L_F(x_N) \\
V^k (x_k) &= \min_{u_k} [L^k(x_k, u_k) + V^{k+1} (f(x_k, u_k))]   \label{bellman eqn} 
\end{align}
where equation \ref{bellman eqn} is the Bellman equation.

In accordance with the right-hand-side term in the Bellman equation, we define the Perturbation as below:

\begin{align} 
\label{perturbation}
P^k (\delta x_k, \delta u_k) &\triangleq L^k (\Tilde{x}_k + \delta x_k, \Tilde{u}_k + \delta u_k) - L^k (\Tilde{x}_k, \Tilde{u}_k) \nonumber \\
&\quad + V^{k+1} (f(\Tilde{x}_k + \delta x_k, \Tilde{u}_k + \delta u_k)) \nonumber \\
&\quad - V^{k+1} (f(\Tilde{x}_k, \Tilde{u}_k))
\end{align}

By expanding it to the second order, we have the following approximation of the Perturbation:
\begin{equation}
P^k (0, 0) = 0
\end{equation}
\begin{align}
\label {taylor series}
P^k (\delta x_k, \delta u_k) &\approx P^k (0, 0) + {P^k_x}^T \delta x_k \nonumber \\
&\quad + {P^k_u}^T \delta u_k \nonumber \\
&\quad + \frac{1}{2} \delta x_k^T P^k_{xx} \delta x_k \nonumber \\
&\quad + \frac{1}{2} \delta u_k^T P^k_{uu} \delta u_k \nonumber \\
&\quad + \delta x_k^T P^k_{xu} \delta u_k + \delta u_k^T P^k_{ux} \delta x_k
\end{align}

From equation \ref{perturbation}, by applying the chain rule, we have the following:
\begin{align}
P^k_x &= L^k_x + f_x^T V_x^{k+1}  \\
P^k_u &= L^k_u + f_u^T V_x^{k+1}  \\
P^k_{xx} &= L^k_{xx} + f_x^T V_{xx}^{k+1} f_x + V_x^{k+1} f_{xx}  \label{tensor1} \\ 
P^k_{uu} &= L^k_{uu} + f_u^T V_{xx}^{k+1} f_u + V_x^{k+1} f_{uu}  \label{tensor2}  \\
P^k_{ux} &= L^k_{uu} + f_u^T V_{xx}^{k+1} f_x + V_x^{k+1} f_{ux}  \label{tensor3}
\end{align}

The last tensor terms in equations \ref{tensor1}, \ref{tensor2}, and \ref{tensor3} can be dropped off, under the linear vehicle model approximation.  

From equation \ref{taylor series}, the approximation of the Perturbation function is convex w.r.t. the action increment. 
\begin{equation} \label{gradient}
\nabla_{\delta u_k} P^k = 0
\end{equation}

Solving equation \ref{gradient}, we have the optimal action increment with respect to the optimal state increment:
\begin{equation} \label{action result}
\delta u_k^{*} = -{P^k_{uu}}^{-1} (P^k_u + P^k_{ux} \delta x_k^{*})
\end{equation}

With the open-loop and feedback terms given as below:
\begin{align} 
q_k \triangleq -{P^k_{uu}}^{-1} P^k_u  \label{open-loop} \\
Q_k \triangleq -{P^k_{uu}}^{-1} P^k_{ux}  \label {feedback}
\end{align}

Plug equation \ref{action result} into \ref{taylor series}, we can represent the optimal k-th step perturbation with respect to $\delta x$:
\begin{equation}
\begin{aligned}
\min_{\delta u_k} P^k(\delta x_k , \delta u_k) &= -\frac{1}{2} {P_u^k}^T {P_{uu}^k}^{-1} P_u^k \nonumber \\
&\quad + ({P_x^k}^T - {P_u^k}^T {P_{uu}^k}^{-1} P_{ux}^k) \delta x_k \nonumber \\
&\quad + \frac{1}{2} \delta x_k^T (P_{xx}^k - {P_{ux}^k}^T {P_{uu}^k}^{-1} P_{ux}^k) \delta x_k 
\end{aligned}
\end{equation}

which yields,
\begin{align}
\Delta V^k = -\frac{1}{2} {P_u^k}^T {P_{uu}^k}^{-1} P_u^k \\
V_x^k = P_x^k - P_{xu}^k {P_{uu}^k}^{-1} P_u^k \\ 
V_{xx}^k = P_{xx}^k - P_{xu}^k {P_{uu}^k}^{-1} P_{ux}^k
\end{align}

Notice that the above steps require to calculate the inverse of the $P_{uu}^k$ matrix. In this paper, we add a regularization term of $10^{-6}$ to each diagonal element of $V_{xx}^{k+1}$ \cite{tassa2012synthesis} to make sure $P_{uu}^k$ is invertible.

\subsubsection{Forward Pass} \hfill\\
Given the open-loop and feedback terms in equation \ref{open-loop} and \ref{feedback}, by applying the vehicle model in a forward order, we can easily get all the updated states and actions:
\begin{align}
x_0 &= x_{start}  \\
u_k &= \Tilde{u}_k + q_k + Q_k (x_i - \Tilde{x}_i)  \\
x_{k+1} &= f(x_k, u_k)
\end{align}

\subsection{Constrained Iterative Linear Quadratic Regulator}

\subsubsection{Constrained Trajectory Optimization Problem}
\begin{equation} \label{constrained traj opt}
\begin{split}
x^{*}, u^{*} = \argmin_{x, u} [{\sum_{0}^{N-1} L^{k} (x_k, u_k)} +L_F(x_N)] \\ 
s.t. \: x_{k+1} = f(x_k, u_k), \: x_0 = x_{start} \\
g^k(x_k, u_k) < 0, g^N(x_N) < 0
\end{split}
\end{equation}

Compared to the Unconstrained Trajectory Optimization Problem described by equation \ref{unconstrained traj opt}, inequality constraints for every state and action, and the final state are considered

The goal of CILQR algorithm is to extend the ILQR algorithm's ability to solve trajectory optimization problem with inequality constraints. 

\subsubsection{Standard Barrier Function} \hfill

The idea is to represent each constraint as an additional cost term, and add it into the corresponding step cost term. We define it as $\beta(g(x, u))$.

The ideal shape of the barrier function is the indicator function, with zero cost when $g(x, u)$ is less than zero and infinite cost elsewhere. However, this makes it not differentiable, thus not solvable by ILQR algorithm. The logarithmic barrier function is chosen to resolve such problem, because it has a similar shape as the indicator function, and also a $C^{\infty}$ function in its domain. 
\begin{equation} \label{log barrier function}
\beta(g(x, u)) = -\frac{1}{t} log(-g(x, u)), g < 0, t > 0
\end{equation}

The greater the parameter $t$ is, the closer between the barrier and the indicator function. Using such standard barrier function requires the initial trajectory must be feasible, such that it satisfies all the inequality constraints.

\subsubsection{Relaxed Barrier Function} \hfill

In order to optimize from any initial trajectory, including infeasible one, we utilize the following Relaxed Barrier Function \cite{hauser2006barrier}:
\begin{equation} \label{relaxed log barrier}
\beta_{relax}(g) = 
    \begin{cases}
        -\frac{1}{t}log(-g) & (g < -\epsilon)\\
        \frac{k-1}{tk}[(\frac{-g - k\epsilon}{(k-1)\epsilon})^k - 1] - \frac{1}{t}log\epsilon & (g \geq -\epsilon)
    \end{cases}
\end{equation}

By choosing $k = 2$, it is guaranteed that the relaxed barrier is a $C^2$ function, for arbitrary $\epsilon$ and $t$ that are greater than 0. In this paper, we choose $t=5$ for all of our inequality constraints. And $\epsilon$ is set accordingly depending on the constraint type.

\subsubsection{Algorithm Structure} \hfill

\textbf{Line Search}: Due to some approximation in the ILQR algorithm (i.e. second order Taylor series, linear vehicle model approximation), it is important to add a line search step after backward pass to further minimize the objective function. In this paper, we use the Golden Section \cite{wright2006numerical} method for line search, which guarantees to find the optimal point for an unimodal objective function within logarithmic time complexity. For general types of objective function, Backtracking \cite{wright2006numerical} method can be used. It keeps multiplying a fixed ratio to $\alpha$ until we have a better solution, with logarithmic time complexity as well. 

\begin{algorithm}
	\caption{CILQR with Relaxed Barrier ($\Tilde{x}, \Tilde{u}, f, L^k, L_F, g, N$)} 
        \label{CILQR_pseudo}
	\begin{algorithmic}[1]
        \State \textbf{Output:} The optimized trajectory ($x^*, u^*$)
		\State \textbf{Repeat}
            \For {$k = 0\;to\;N-1$}
            \State $L^k \leftarrow L^k + \beta_{relax}(g^k(., .))$
            \EndFor
            \State \quad $L_F \leftarrow L_F + \beta_{relax}(g^N(.))$
            \For {$k = N-1\;to\;0$}
            \State Compute $q_k, Q_k$ via \textbf{Backward Pass}
            \EndFor
            \State \quad Compute \textbf{$\alpha$} $(0 < \alpha <= 1)$ that minimize the objective function described in equation \ref{constrained traj opt} via \textbf{Line Search}
            \State \quad $\Hat{x}, \Hat{u} \leftarrow \Tilde{x}, \Tilde{u}$
            \For {$k = 0\;to\;N-1$} 
            \State $\Hat{u}_k \leftarrow \Hat{u}_k + \alpha * (q_k + Q_k(\Hat{x}_k - \Tilde{x}_k))$
            \State $\Hat{x}_{k+1} = f(\Hat{x}_k, \Hat{u}_k)$
            \EndFor
			\State \quad $\Tilde{x}, \Tilde{u} \leftarrow \Hat{x}, \Hat{u}$
		\State \textbf{until} converge
        \State $x^*, u^* \leftarrow \Tilde{x}, \Tilde{u}$ 
	\end{algorithmic} 
\end{algorithm}

We set the converging condition based on the percentage of the cost drop between two adjacent iterations: with a threshold of $1\%$.  

Combining backward pass, forward pass, barrier function, and line search techniques, we have the complete CILQR algorithm structure shown in Algorithm \ref{CILQR_pseudo}.

\section{PROBLEM FORMULATION}

\subsection{Vehicle Model}

One originality of this paper comes from the adoption of higher-order control terms in the vehicle model. For comparisons, \cite{pan2020safe} used the curvature terms as the lateral control input for the kinetic vehicle model. The curvature term could be directly related to the steering of the vehicle by: 
\begin{equation} \label{eqn kappa}
\kappa = \frac{ \tan \delta}{L} ,
\end{equation}
based on the kinematic model, where $\delta$, $L$ is the front steering angle and the wheelbase length, respectively. 

The afterward ILQR problem formation therefore does not bring the higher-order terms of the curvature into consideration and may lead to sudden change of the curvature in the afterward planning. This would be unfavorable for the driving comfort experience since the passengers could feel the exceedingly large lateral jerk caused by the large steering rate: 
\begin{equation} \label{lat jerk equation}
j_y = \frac{\mathrm{d} }{\mathrm{d} t}\left ( v^2\kappa \right )  = 2v\kappa a_x + \frac{v^2}{L}\sec^2 (\delta) \dot \delta = 2v\kappa a_x + v^2\dot{\kappa}
\end{equation}

From the above equation, it is obvious that the steer rate $\dot \delta$, or equivalently $\dot \kappa$, could directly influence the lateral jerk. 

Moreover, to ensure the executability of the trajectory by the controller, the steer rate planned should also remain within limits of the steering actuators during planning or otherwise could cause vehicle deviation from the trajectory.

For these reasons, we replace steering with the rate of the curvature as the lateral control input for the benefits of controllable guarantees and enhanced driving comfort.

In this paper, we use vehicle kinematic model with higher order terms, for controllable guarantee 
 and enhanced comfort. The state vector is defined as $[x, y, v, \theta, a, \kappa, \Dot{\kappa}]^T$, where $(x, y, \theta)$ describes APV's 2D position and heading angle in Cartesian frame, v is the longitudinal speed, a is the longitudinal acceleration, $\kappa$ and $\Dot{\kappa}$ are curvature and its first derivative w.r.t. time, respectively. The action vector is defined as $[j, \Ddot{\kappa}]^T$, where $j$ is the longitudinal jerk, and $\Ddot{\kappa}$ is the curvature's second derivative w.r.t. time. Between two adjacent states, the time step is $T_r$, and we assume constant $j$ and $\Ddot{\kappa}$. The distance step is given by $S_r = v_0 T_r + \frac{1}{2} a_0 T_r^2 + \frac{1}{6} j_0 T_r^3$, and the average curvature is given by $\bar \kappa = \frac{1}{T_r} \int_{0}^{T_r} \kappa (t) dt = \kappa_0 + \frac{1}{2} \dot{\kappa}_0 T_r + \frac{1}{6} \Ddot{\kappa}_0 T_r^2$. We assume a constant derivative of heading angle w.r.t. travel distance (i.e. $\frac{d \theta}{ds} = \bar \kappa$), when updating the 2D position.

 The updated states are given as below:
\begin{equation} \label{state x}
\begin{aligned}
x_1 &= x_0 + \int_{0}^{S_r} \cos{(\theta_0 + \bar \kappa s)}ds \\
&= x_0 + \frac{\sin{(\theta_0 + \bar \kappa S_r)} - \sin{\theta_0}}{\bar \kappa} (\bar \kappa \neq 0) \\
&= x_0 + S_r \cos{\theta_0} (\bar \kappa = 0)
\end{aligned}
\end{equation}
\begin{equation} \label{state y}
\begin{aligned} 
y_1 &= y_0 + \int_{0}^{S_r} \sin{(\theta_0 + \bar \kappa s)}ds \\ 
&= y_0 + \frac{-\cos{(\theta_0 + \bar \kappa S_r)} + \cos{\theta_0}}{\bar \kappa} (\bar \kappa \neq 0) \\
&= y_0 + S_r \sin{\theta_0} (\bar \kappa = 0)
\end{aligned}
\end{equation}
\begin{align} 
v_1 &= v_0 + a_0 T_r + \frac{1}{2} j_0 T_r^2  \label{state v} \\ 
\theta_1 &= \theta_0 + \int_{0}^{S_r} \kappa (s) ds = \theta_0 + \int_{0}^{T_r} \kappa (t) v(t) dt  \label{state theta} \\
\kappa_1 &= \kappa_0 + \dot \kappa_0 T_r + \frac{1}{2} \ddot \kappa_0 T_r^2 \label{state kappa} \\
\dot \kappa_1 &= \dot \kappa_0 + \ddot \kappa_0 T_r  \label{state kappa dot}
\end{align}

By combining equation \ref{state x}, \ref{state y}, \ref{state v}, \ref{state theta}, \ref{state kappa}, and \ref{state kappa dot}, we have the complete vehicle model as below:
\begin{equation} \label{vehicle model}
\begin{bmatrix} x_{k+1} \\ y_{k+1} \\ v_{k+1} \\ \theta_{k+1} \\ a_{k+1} \\ \kappa_{k+1} \\ \dot{\kappa}_{k+1} \end{bmatrix} = f(\begin{bmatrix}
    x_k \\ y_k \\ v_k \\ \theta_k \\ a_k \\ \kappa_k \\ \dot{\kappa}_k 
\end{bmatrix}, 
\begin{bmatrix}
    j_k \\ \Ddot{\kappa}_k 
\end{bmatrix})
\end{equation}

\subsection{Cost Function} \label{cost function section}

For CILQR algorithm, there are two categories of costs, namely state and action costs. In the previous work \cite{8671755}, four types of state costs are considered, including acceleration, curvature, velocity tracking, and reference tracking. The first two terms are quadratic w.r.t. the absolute magnitude, while the last two terms are quadratic w.r.t. the difference with their references, respectively. 

Based on our higher-order vehicle model described in equation \ref{vehicle model}, we introduce the following additional state and action costs, to enhance the overall comfort and controllability. In general, quadratic function is preferred due to its convexity and symmetry.

\subsubsection{Lateral Acceleration} \hfill

Given no side slip of our vehicle kinematic model, the lateral acceleration is given as below:
\begin{equation} \label{lat accel}
    a_y = \kappa v^2
\end{equation}

where it can be directly expressed by current states. Then we assign lateral acceleration a quadratic cost, which penalizes rapid change in heading angle, with significantly more penalty in high speed compared to low speed.
\begin{equation}
    cost_k^{a_y} = w_{a_y} \cdot {a_y}_k^2
\end{equation}


\subsubsection{Lateral Jerk} \hfill

The expression of lateral jerk is given by equation \ref{lat jerk equation}, where it can also be directly expressed by current states. Then we assign lateral jerk a quadratic cost, which penalizes rapid change in lateral acceleration
\begin{equation}
    cost_k^{j_y} = w_{j_y} \cdot {j_y}_k^2
\end{equation}


\subsubsection{Longitudinal Jerk} \hfill

Since longitudinal jerk is one of the existing action terms, we can directly assign it a quadratic cost, to penalize rapid change in longitudinal acceleration
\begin{equation}
    cost_k^{j} = w_{j} \cdot j_k^2
\end{equation}


\subsubsection{Direction Tracking} \hfill

For on-road trajectory planning, it is important to encourage APV to keep its heading angle aligned with the road's forward direction. Therefore we introduce a quadratic cost w.r.t. the heading error
\begin{equation}
    cost_k^{\theta} = w_{\theta} \cdot (\theta_k - \theta_k^r)^2
\end{equation}

\subsection{Constraints}

There are three categories of constraints we need to consider, namely, kinematic, roadway-compliance, and collision avoidance

\subsubsection{Kinematic Constraints} \hfill

For drive-by-wire autonomous vehicles, the lower and upper bounds of steer angle, steer rate, steer acceleration, and longitudinal acceleration, are limited due to the underneath vehicle actuation mechanism. Therefore, we can get those limits via system identification. In the previous work \cite{8671755}, steer angle and steer rate constraints are considered. 

Besides those constraints, we apply constraints to the first and second derivative of curvature w.r.t. time, by taking the derivative on both sides of equation \ref{eqn kappa}:
\begin{align} 
\dot{\kappa} &= \frac{1 + (\kappa L)^2}{L} \cdot \dot{\delta}  \label{eqn kappa dot} \\
\Ddot{\kappa} &= \frac{1 + (\kappa L)^2}{L} \cdot \Ddot{\delta} + \frac{2 \kappa (1 + (\kappa L)^2)}{L} \cdot \dot{\delta}^2  \label{eqn kappa dot dot}
\end{align}

where $\dot{\delta}$ and $\Ddot{\delta}$ are the steer rate and steer acceleration, respectively, and $\kappa$ is APV's current curvature. From equation \ref{eqn kappa dot}, and \ref{eqn kappa dot dot}, given the limits of $\dot{\delta}$ and $\Ddot{\delta}$, we can calculate the corresponding limits of $\dot{\kappa}$ and $\Ddot{\kappa}$, and then apply constraint costs onto them via the relaxed barrier in equation \ref{relaxed log barrier}.

\subsubsection{Roadway-Compliance Constraints} \hfill



In on-road trajectory planning, it's crucial that the APV does not cross non-traversable lane boundaries, such as solid lines or curbs. We model the APV as a 2D rectangular bounding box and the lane boundary as a polyline comprised of multiple line segments. To enforce the lane boundary constraint, we compute the distance between the bounding box and the polyline, ensuring it remains above a predetermined lower limit through the application of a relaxed barrier function. Additionally, we establish a distance threshold of 1.0 meters. If this threshold is breached, a quadratic soft cost is imposed for the extent of the violation.

\subsubsection{Collision Avoidance Constraints} \hfill


Collision avoidance constraints are applied between the APV and surrounding obstacles. Both the APV and obstacles are represented as 2D bounding boxes. For each pair, the box-to-box distance is calculated to ensure that it exceeds a specified lower limit. This is enforced through the application of a relaxed barrier function.

\section{CASE STUDY}

The proposed modified CILQR algorithm has been realized in C++14 and is running in a single thread under Intel(R) Core(TM) i7-10870H CPU @ 2.20GHz. The algorithm is then deployed and validated both in simulation and on actual road tests with interacting traffic. The average run-time during the run is around 20 ms with a standard deviation 5 ms, satisfying the regular need of planning cycle within 100 ms. For the calculation of Jacobian and Hessian matrices, since we are mainly using Numerical Differentiation, which has a quadratic time complexity w.r.t. the number of states or actions, the run-time can be further optimized with Automatic Differentiation \cite{baydin2018automatic}.

\begin{table}[h]
\centering
\caption{APV Specification}
\label{table_example}
\begin{tabular}{cc}
\toprule
\textbf{Parameter} & \textbf{Value} \\
\midrule
Length [$m$] & 4.77 \\
Width [$m$] & 1.93 \\
Max Accel [$m/s^2$] & 5.0 \\
Max Decel [$m/s^2$] & -5.0 \\
Max Steer Angle [$^\circ$] & 475.0 \\
Max Steer Rate [$^\circ/s$] & 550.0 \\
Max Steer Acceleration [$^\circ/s^2$] & 1200.0 \\
Steer Ratio & 15.8 \\
Wheel Base [$m$] & 2.88 \\
\bottomrule
\end{tabular}
\end{table}

\includegraphics[width=1.0\linewidth]{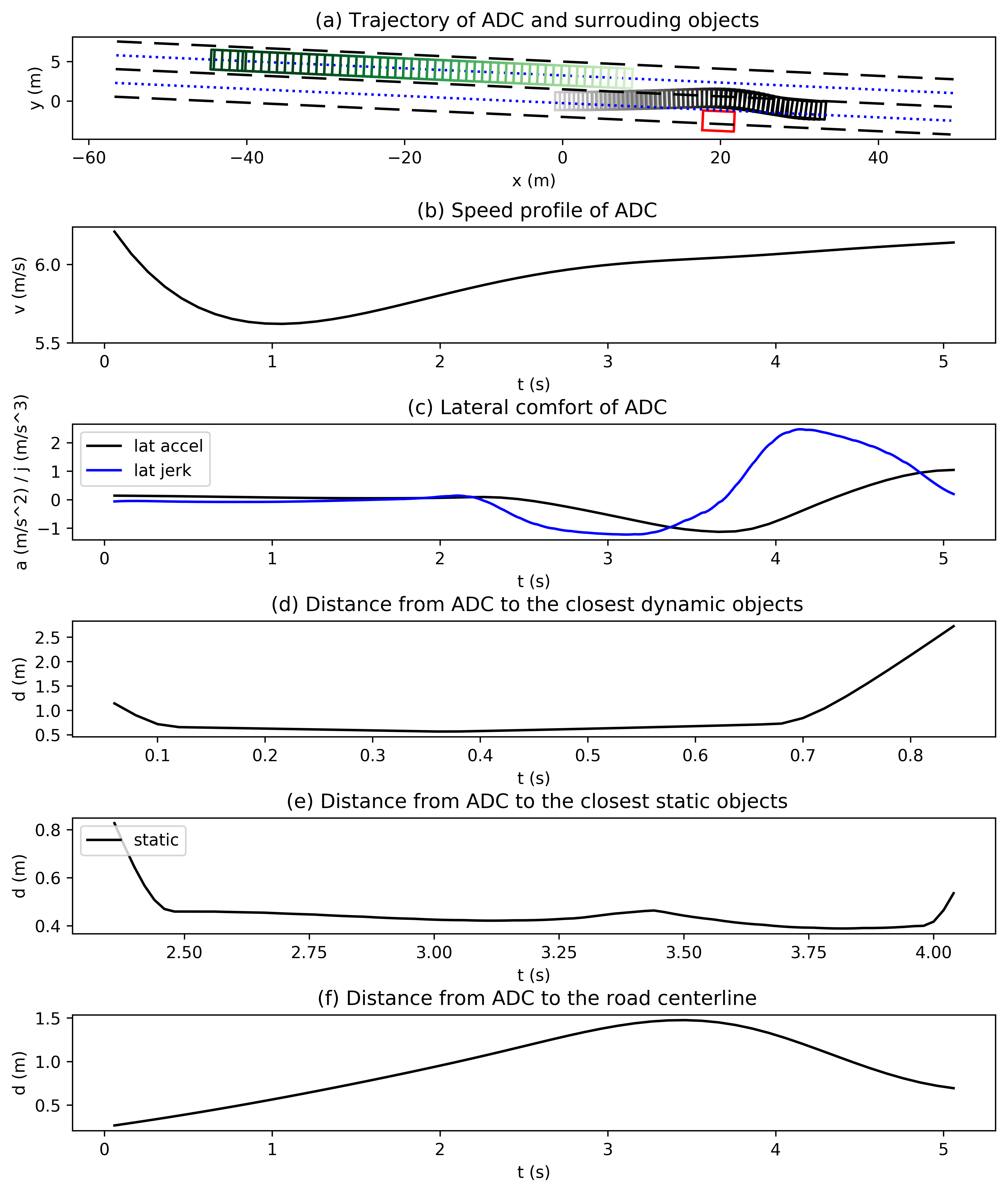} 
\captionof{figure}{Overtake Street Parking Vehicles Borrowing Opposite Lane}\label{static avoidance png}
From the extensive test data, three scenarios are chosen to demonstrate our method's capability of balancing comfort, roadway-compliance, and safety. For each trajectory point, we also plot its longitudinal and lateral physical quantities, as well as obstacle and roadway-compliance related distance terms. The planning horizon is set to 50 time steps with the discrete time step 0.1 second. The passenger vehicle we use for the demonstration have the specifications listed in Table \ref{table_example}. The planned trajectory is then executed with decoupled longitudinal and lateral controllers similar to \cite{xu2018accurate}\cite{zhu2019parameter}\cite{xu2019design} running at 50 Hz. 

\subsection{Overtake Street Parking Vehicles Borrowing Opposite Lane}
In the simulation depicted in Fig \ref{static avoidance png}, the APV begins centered in its lane, moving at a speed of 10 $m/s$, as displayed in (b). The lane itself is 3.5 meters wide. Ahead of the APV, at a distance of 30 meters, there's a street-parked vehicle occupying 1.7 meters of the lane. Concurrently, a vehicle is traveling in the opposite lane at a speed of 10 $m/s$. The APV's objective is to safely and smoothly pass the parked vehicle without coming too close to the oncoming vehicle. As seen in (a) and (b), the APV decelerates to make way for the oncoming vehicle while simultaneously executing lateral adjustments. The maneuvers ensure comfort metrics, such as lateral acceleration and jerk, remain within acceptable ranges, as highlighted in (c). Furthermore, the APV consistently maintains a minimum distance of 0.4 meters from both the moving and parked vehicles. It also aligns closely with the lane's forward direction and stays near the lane's center, as illustrated in (d), (e), and (f).

\subsection{Dynamic Cut-in Obstacle Collision Avoidance}
In the road test depicted in Fig \ref{dynamic avoidance png}, the APV starts off centered in its lane, traveling at an approximate speed of 12 m/s, as illustrated in (b). Directly ahead of the APV, another vehicle cuts in from the right at a similar speed, initially just about 2 meters away, as displayed in (d). This presents a significant collision risk. To ensure safe following and lateral distances, the APV decelerates slightly and makes a subtle in-lane shift to the left. Furthermore, the APV manages to maintain all comfort metrics within the standard parameters, as highlighted in (c). It also stays close to the lane's center, deviating by no more than 0.4 meters laterally, as indicated in (e).

\subsection{Road Edge Avoidance on Freeway}
During the freeway road test depicted in Fig \ref{road edge avoidance png}, the APV starts nearly centered in its lane, moving at a speed close to 28 $m/s$, as illustrated in (b). To its left, there's a road edge, deemed as an non-crossable lane boundary. Initially, this boundary is less than 0.6 meters from the APV, as presented in (d). The APV's objective, which it meets successfully, is to smoothly shift away from the road edge, maintaining a safe lateral distance while staying close to the lane's center. By the end of its trajectory, the distance from the road edge extends to 1.0 meter, with the maximum lateral deviation from the lane center being about 0.4 meters, as indicated in (d) and (e). The top readings for lateral acceleration and jerk are approximately 1 $m/s^2$ and 2 $m/s^3$, respectively, as highlighted in (c), figures that align with typical human driving data.

\section{CONCLUSIONS}
In this paper, we've successfully integrated higher-order dynamics and roadway-compliance into the CILQR algorithm, as a trajectory planner for autonomous vehicle. Higher-order dynamics, which is crucial for comfort and controllability,  include expanding additional states, such as the first and second derivatives of curvatures and longitudinal jerk, and penalize additional comfort metrics such as lateral acceleration and jerk. Roadway-compliance includes encouraging APV to perform reference and direction tracking, and maintain a proper distance to the non-crossable lane boundaries. Further, we successfully implemented the algorithm in real-time simulation and road tests with interacting traffic. The case study results show that our improved CILQR trajectory planner is capable of balancing comfort, roadway-compliance, and safety based on our cost and constraint design. 

\includegraphics[width=1.0\linewidth]{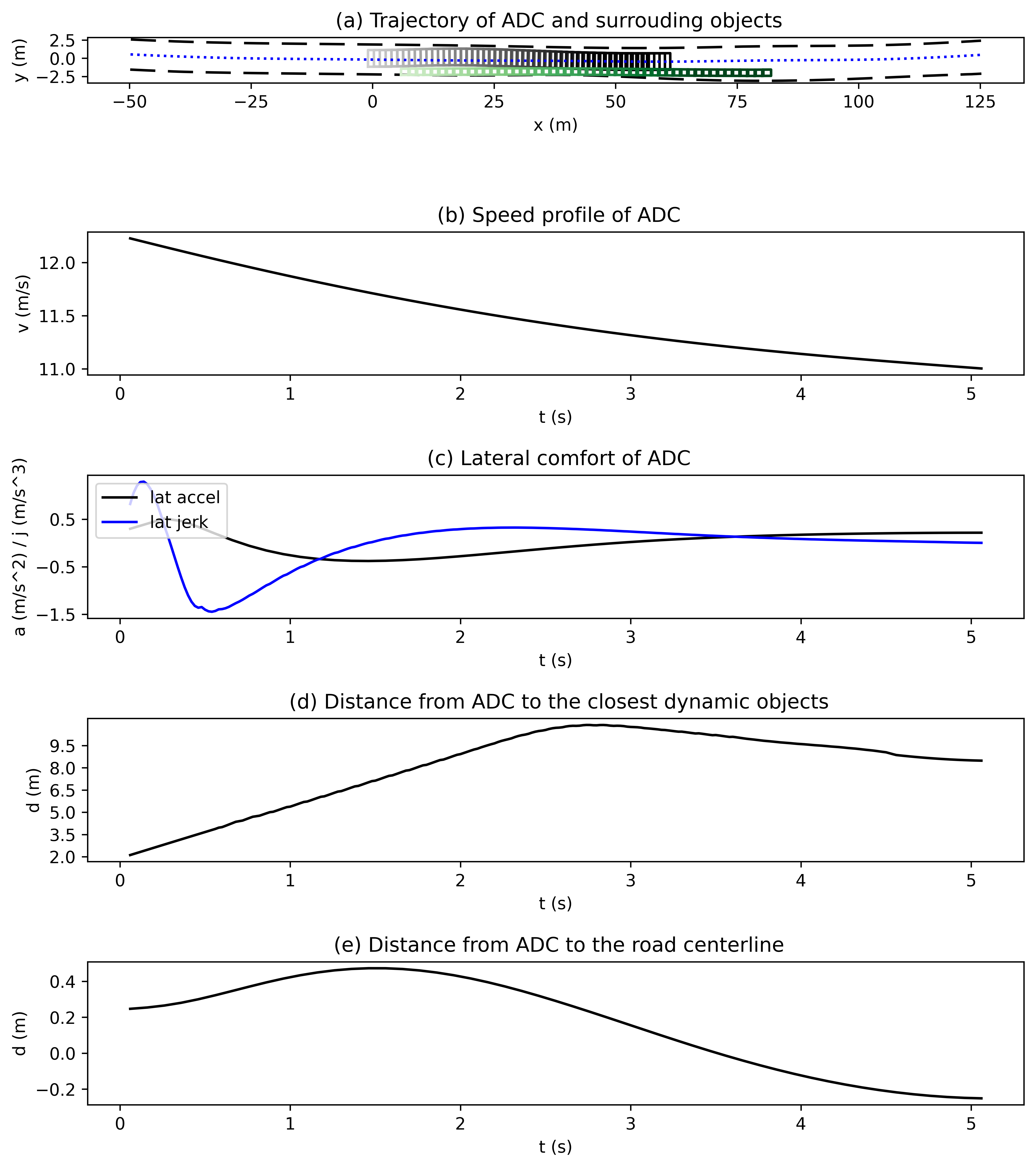} 
\captionof{figure}{Dynamic Cut-in Obstacle Collision Avoidance (road test)}\label{dynamic avoidance png}

\includegraphics[width=1\linewidth]{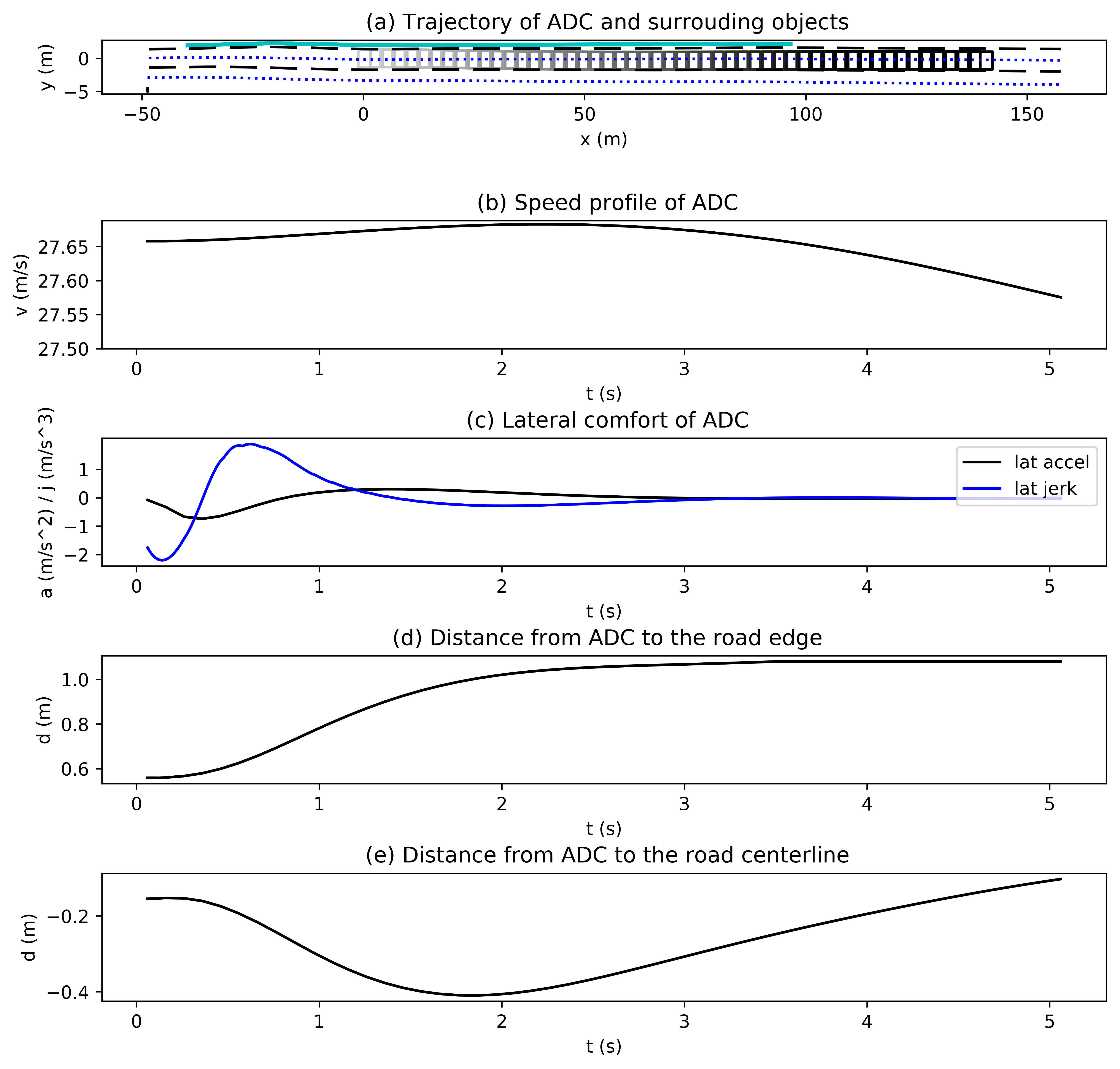} 
\captionof{figure}{Road Edge Avoidance on Freeway (road test)} \label{road edge avoidance png}
\addtolength{\textheight}{-12cm}   





\bibliographystyle{IEEEtran}
\bibliography{ref}




\end{document}